%% file: main.tex
\documentclass{article} 
\usepackage{iclr2020_conference,times}

\input{math_commands.tex}

\usepackage{hyperref}
\usepackage{url}
\usepackage{amsmath, mathtools}
\usepackage{array, url, subcaption}

\usepackage{dblfloatfix} 

\usepackage{times} 

\iclrfinalcopy
\begin{document}

\title{Using competency questions to select optimal clustering structures for residential energy consumption patterns}

\author{Wiebke Toussaint \thanks{Work done while at the Department of Computer Science at the University of Cape Town} \\
Department of Technology, Policy \& Management\\
Delft University of Technology\\
Netherlands \\
\texttt{w.toussaint@tudelft.nl} \\
\And
Deshendran Moodley \\
Department of Computer Science \\
University of Cape Town \& \\  
Centre for Artificial Intelligence Research\\
South Africa \\
\texttt{deshen@cs.uct.ac.za} \\
}

\maketitle

\begin{abstract}
During cluster analysis domain experts and visual analysis are frequently relied on to identify the optimal clustering structure. This process tends to be adhoc, subjective and difficult to reproduce. This work shows how competency questions can be used to formalise expert knowledge and application requirements for context specific evaluation of a clustering application in the residential energy consumption sector. 

\end{abstract}

\input{sections/previous_work}
\input{sections/data}
\input{sections/evaluation_framework}
\input{sections/results}
\input{sections/discussion}

\subsubsection*{Acknowledgements}
This research was funded in part by the South African Centre for Artificial Intelligence Research (CAIR).

{\footnotesize
\bibliography{library}}
\bibliographystyle{iclr2020_conference}

\appendix
\input{sections/appendix_a}
\input{sections/appendix_b}
\input{sections/appendix_c}

\end{document}

%% file: math_commands.tex
\usepackage{amsmath,amsfonts,bm}

\def\eqref#1{equation~\ref{#1}}

\def\1{\bm{1}}

\DeclareMathAlphabet{\mathsfit}{\encodingdefault}{\sfdefault}{m}{sl}
\SetMathAlphabet{\mathsfit}{bold}{\encodingdefault}{\sfdefault}{bx}{n}



%% file: sections/previous_work.tex
\section{Background and Previous Work}
\label{previous_work}

While cluster analysis is an established unsupervised machine learning technique, identifying the optimal set of clusters for a specific application requires extensive experimentation and domain knowledge. Cluster compactness and distinctness are two important attributes that characterise a good cluster set \citep{Sarle1990} and different metrics, such as the Mean Index Adequacy (MIA), Davies-Bouldin Index (DBI) and the Silhouette Index have been proposed to measure cluster compactness and distinctness. In practise, a combination of measures together with additional expert guidance and visual inspection of clustering results is often used during the experimental process to identify the best cluster set \citep{Jin2017}, \citep{Dent2014a}. However, these qualitative approaches can be adhoc and time consuming, subjective and difficult to reproduce, and biased by the expert's interpretation of the visual representation \citep{Gogolou2019}. This work shows how competency questions from the ontology engineering community can be used to guide cluster set selection for generating representative daily load profiles that are suitable for developing customer archetypes of residential consumers in South Africa.

A daily load profile describes the energy consumption pattern of a household over a 24 hour period. Representative daily load profiles (RDLPs) are indicative of distinct daily energy usage behaviour for different types of households. Customer archetypes are developed to represent groupings of energy users that consume energy in a similar manner. RDLPs have been well explored for generating customer archetypes for applications in long term energy modelling \citep{Figueiredo2005}, \citep{McLoughlin2015}. Traditionally, the most common approaches used for clustering load profiles are centroid-based approaches and variants of kmeans, self-organising maps (SOM) and hierarchical clustering \citep{Jin2017}. For residential consumers the variable nature of individual households makes the interpretation of clustering results ambiguous \citep{Swan2009}, a challenge that is exacerbated in highly diverse, developing country populations, where economic volatility, income inequality, geographic and social diversity contribute to increased variability of residential energy demand \citep{Heunis2014}. \citet{Xu2017} have used pre-binning, which involves applying a two-stage clustering algorithm that first clusters load profiles by overall consumption and then by load shape, to improve clustering results for highly variable households spread across the United States. In addition to the general clustering metrics, \citet{Kwac2014} also propose the notion of entropy as a metric for capturing the variability of electricity consumption of a household. To evaluate the result of segmenting a large number of daily load profiles into interpretable consumption patterns, \citet{Xu2017} use peak overlap, percentage error in overall consumption and entropy as metrics.

In ontology engineering, competency questions are an established methodology used to specify the requirements of an ontology and to evaluate the extent to which a particular ontology meets these requirements \citep{Gruninger1995}. Brainstorming, expert interviews and consultation of established sources of domain knowledge are processes that can be used to identify competency questions \citep{denicola2009}. Informal competency questions can be expressed in natural language and connect a proposed ontology to its application scenarios, thus providing an informal justification for the ontology \citep{Uschold96ontologies}. To our knowledge competency questions have not been used previously to evaluate clustering structures.

%% file: sections/data.tex
\section{Data}
\label{data}

The Domestic Electrical Load Metering Hourly (DELMH) \citep{delmh2019} dataset contains 3~295~194 daily load profiles for 14~945 South African households over a period of 20 years from 1994 to 2014. The daily load profile $h_d^{(j)}$ is a 24 element vector $l(t)$ representing the hourly consumption (measured in Amperes) of household $j$ on day $d$. Each interval $t$ is labeled by the start time, such that $t = 0$ captures interval 00:00:00 - 00:59:59. $H^{(j)}$ is the array of all daily load profile vectors $h^{(j)}$ for household $j$, and $X$ (dim 3 295 848 $\times$ 24) is the array of all daily load profiles $h$.

\begin{equation}
    h_d^{(j)} = l(t) \textrm{, where } t=\{0, 1 ... 23\} 
\end{equation}
\ignorespacesafterend
\begin{equation}
    H^{(j)} = \big[h_d^{(j)}\big] \textrm{, where }  d=\{1, 2 ... d \textrm{ days}\}
\end{equation}
\ignorespacesafterend
\begin{equation}
    X = \big[H^{(j)}\big] \textrm{, where } j=\{1, 2 ... 14945\} 
\end{equation}

We can then use clustering to find an optimal clustering structure $k$, given the dataset $X$.

%% file: sections/evaluation_framework.tex
\section{Developing Competency Questions}
\label{qualitative_evaluation}

We used a combination of analysing existing standards and engagement with domain experts to formulate informal competency questions expressed in natural language. The Geo-based Load Forecasting Standard \citep{GLF2012} contains manually constructed load profiles and guiding principles for load forecasting in South Africa. The competency questions were developed after analysis of this standard and continuous engagement with a panel of five industry experts. There were initial interviews with all experts to elicit the usage requirements. Preliminary competency questions were presented at a workshop with key stakeholders in the community. The final version of the competency questions incorporated the feedback from the stakeholders. The competency questions were then used to construct associated  qualitative evaluation measures and a cluster scoring matrix that weights these measures to provide a qualitative ranking of cluster sets in terms of the application requirements. 

The following five core competency questions were identified: 
\begin{enumerate}
\itemsep0em 
    \item Can the load shape and demand be deduced from clusters?
    \item Do clusters distinguish between low, medium and high demand consumers?
    \item Can clusters represent specific loading conditions for different day types and seasons?
    \item Can a zero-consumption profile be represented in the cluster set\footnote{This was deemed important for considering energy access in low income contexts, as households may go through periods where they cannot afford to buy electricity and thus have no consumption.}?
    \item Is the number of households assigned to clusters reasonable, given knowledge of the sample population?
\end{enumerate}

Based on these questions, we defined a good cluster set as having expressive clusters and being usable. An expressive cluster must convey specific information related to particular socio-economic and temporal energy consumption behaviour. A usable cluster set must represent energy consumption behaviour that makes sense in relation to the clustering context and that carries the necessary information to make it pertinent to domain users. Next we developed qualitative measures to assess the competency questions. They are explained briefly below and in detail in Appendix A.

Expressivity (from competency questions 2 and 3) requires that the RDLP of a cluster is representative of the energy consumption behaviour of the individual daily load profiles that are members of that cluster, as expressed by the mean consumption error of total and peak demand and the mean peak coincidence ratio. An expressive cluster must also have the ability to convey specific meaning, especially in contexts where populations are highly variable. Cluster entropy can be used as a measure to establish the information embedded in a cluster and thus its specificity. The lower the entropy, the more information is embedded in the cluster, the more specific (homogeneous) the cluster, the better the cluster. In a specific cluster all members share the same context, e.g. daily load profiles of low consumption households on Sundays in summer.

The characteristic of cluster usability was derived from competency questions 4 and 5. Question 4 requires a manual evaluation based on expert judgement and is evaluated as being either true, or false. Question 5 is calculated as the percentage of clusters whose membership exceeds a threshold value of 10490 members\footnote{The threshold was selected as a value approximately equal to 5\% of households using a particular cluster for 14 days.}. Additional considerations are that fewer clusters typically ease interpretation and are thus preferable to larger numbers of clusters. The maximum number of clusters should be limited to 220, based on population diversity and existing expert models which account for 11 socio-demographic groups, 2 seasons, 2 daytypes and 5 climatic zones.

\subsection{Cluster Scoring Matrix}
The cluster scoring matrix in Table \ref{tab:cluster_scoring_matrix} presents a summary of the attributes and competency questions, the corresponding evaluation measures and their weights. The weights are based on the relative importance that experts assigned to the measure. Experiments are ranked by performance in each measure, with a score of 1 indicating the best cluster set. A weighted score is then computed for each experiment by multiplying its rank with the corresponding measure’s weight, and summing over all measures. The lower the total score, the better the cluster set.

\begin{table}[h]
{\caption{Cluster Scoring Matrix}\label{tab:cluster_scoring_matrix}}
\begin{center}
\scriptsize
\begin{tabular}{p{0.14\linewidth} p{0.07\linewidth} p{0.32\linewidth} c c}
    \hline\noalign{\smallskip}
    \textbf{Attribute} & \textbf{Qu.} & \textbf{Evaluation measure} & & \textbf{Weight} \\
    \noalign{\smallskip}\hline\noalign{\smallskip}
     usable  & 5& sensible count per cluster & & 2\\ \noalign{\smallskip}
      & 4 & zero-profile representation & & 1\\ \noalign{\smallskip}\hline\noalign{\smallskip}
     expressive & 1 & mean consumption error & total & 6\\
     representative & 1 & & peak & 6\\
      & 1 & mean peak coincidence & & 3\\ \noalign{\medskip}
    expressive & 3 & temporal entropy & weekday & 4\\ 
    specific & 3 & & monthly & 4\\
      & 2 & demand entropy & total daily & 5\\
      & 2 & & peak daily & 5\\
\end{tabular}
\end{center}
\end{table}
\vspace{-1em}

%% file: sections/results.tex
\section{Clustering Experiments and Results}
\label{qual_results}

Various clustering experiments were performed to find a set of clusters $k^{(i)}$ that symbolise the best RDLPs $R^{(i)}$ for $X$. The clustering process was set up as a typical data processing pipeline, using hourly daily load profiles from DELMH as input. Depending on the experiment, different pre-processing steps were performed. These include the selection of a pre-binning by average monthly consumption (AMC) or integral k-means, and retaining or dropping zero values. Each of the experiments was run with four different normalisation algorithms, and without  normalisation. Algorithms were initialised with different parameter values to generate cluster sets with a range of membership sizes. Details on the algorithms, normalisation and pre-binning are provided in Appendix B. 

\subsection{Evaluation}
Based on the experiment details defined in Table \ref{tab:experiments} in  Appendix \ref{appendixb}, 2083 individual experiment runs were conducted across all parameters. Each run was first evaluated with traditional quantitative clustering metrics. To ease the quantitative evaluation process and allow for comparison across metrics, Mean Index Adequacy (MIA), the Davies-Bouldin Index (DBI) and Silhouette Index were combined into a Combined Index (CI) score. The top 10 ranked experiment runs based on the CI score are shown in Table \ref{tab:clusters_top10} in Appendix \ref{appendixc}. The highest ranked experiments were then further evaluated with the cluster scoring matrix.

\subsection{Qualitative Clustering Results}

Table \ref{tab:qual_ranked_results} in Appendix \ref{appendixc} summarises the scores and ranking produced by the cluster scoring matrix. The scores span a greater range of values than the CI scores and are grounded in interpretable measures, which makes the results more meaningful and eases the selection of the best experiment. While the top two runs lie only 8 points apart, they comfortably outperform the third best run, which has double the score. The potential of the qualitative evaluation measures is evident when contrasting the quantitative and qualitative results of exp. 5 (kmeans, zero-one) with those of exp. 8 (kmeans, unit norm). Exp. 5 (kmeans, zero-one) had the second best run based on the CI score but was ranked second last in the cluster scoring matrix. Exp. 8 (kmeans, unit norm) on the other hand only ranked ninth by quantitative score, yet convincingly claimed the top position based on qualitative measures. 

Comparing the RDLPs in Figure \ref{fig:cluster_comparison} in Appendix \ref{appendixc} gives confidence in the reranking. Exp. 5 (kmeans, zero-one) has only 18 clusters; on average 2.125 clusters per bin. The five smallest clusters combined have fewer than 1500 member profiles and appear invisible in the bar chart of cluster size at the bottom of Figure \ref{fig:cluster_centroids_exp5_kmeans_zero-one}. The ragged shapes of \textit{cluster 16}, \textit{cluster 17} and \textit{cluster 18} are also an indication that very few profiles were aggregated in these RDLPs. Over half of all load profiles belong to only three clusters: \textit{cluster 5}, \textit{cluster 6} and \textit{cluster 9}. As a whole, the individual RDLPs lack distinguishing features and are neither expressive nor useful, making them poorly suited for creating customer archetypes.

Exp 8 (kmeans, unit norm) on the other hand has 59 clusters, varying between 2 and 15 clusters per bin. With the exception of \textit{cluster 33} which accounts for roughly 15\% of all daily load profiles, cluster membership for the remaining clusters varies in a range from 15 000 to 100 000 members. \textit{Cluster 33} is one of only two clusters in its bin, which has a large bin membership in line with expectations given our sample population. Collectively, the individual RDLPs are expressive, featured and distinct, which promises that they will be useful for constructing customer archetypes.

%% file: sections/discussion.tex
\section{Discussion and Conclusion}
\label{discussion}

This work formalises competency questions, formulated in consultation with domain experts, as quantifiable, qualitative evaluation measures. The qualitative measures are summarised in a cluster scoring matrix which weights, ranks and compares the measures across clustering experiments. By combining traditional clustering metrics and qualitative evaluation measures, clustering structures with good compactness and distinctness are thus ranked by their usability and expressivity, which guides our selection of a clustering structure that is useful for our intended application of creating customer archetypes in the residential energy sector in South Africa.

The cluster scoring matrix eases the scoring and ranking of experiments, while also making the reliance on expert validation explicit and repeatable. It clearly indicates that of the top 10 experiments, unit norm normalisation and pre-binning produced the most expressive and usable clusters. While the best experiment was pre-binned with integral kmeans, pre-binning by average monthly consumption produced comparable scores. The difference in scores between the two pre-binning approaches was strongly influenced by the weights assigned to different evaluation measures and the threshold determining the minimum cluster membership. These are subjective constraints determined by our application context. In a different application, they may be set differently. The cluster scoring matrix could be improved by making it less susceptible to weight and threshold changes, as well as the ranking method. A limitation of the work is that we used well established clustering techniques and have not tested more recent clustering algorithms and dynamic time warping. 

Our work presents a novel application of machine learning in the energy domain in South Africa, with potential for application in other developing country contexts. The approach shows promise for generating clusters that are useful for application in a real-world, long-term energy planning scenario and demonstrates the use of cluster analysis techniques for building real world systems.

%% file: sections/appendix_a.tex
\section{Qualitative Evaluation Measures}
\label{appendixa}

We use $k_x$ to denote a single cluster in clustering structure $k$. The score $S^m$ of a qualitative measure for cluster set $k$ is the mean of the scores $S^m_x$ of all clusters $k_x$ with more than 10490 members. Clusters with a small member size were excluded when calculating mean measures, as they tend to overestimate the performance of poor clusters. Individual cluster performance is weighted by cluster size to account for the overall effect that a particular cluster has on the set.

\subsection{Mean Consumption Error}
\label{mean_consumption_error}

The total daily demand and peak daily demand for an actual daily load profile $l(t)_d$ and a predicted cluster representative daily load profile $l(t)_x$ are given by the equations below: 

\begin{eqnarray}
    d^{(j)}_{d_{total}} = \sum_{t=0}^{23}{l(t)_d} &\textrm{ and }& d^{(j)}_{d_{peak}} = l(t)^{max}_d\\
    d^{(R)}_{x_{total}} = \sum_{t=0}^{23}{l(t)_x} &\textrm{ and }& 	d^{(R)}_{x_{peak}} = l(t)^{max}_x
\end{eqnarray}

Four mean error metrics are calculated to characterise the extent of deviation between the total and peak demand of a cluster, and those of its member profiles. Mean absolute percentage error (MAPE) and median absolute percentage error (MdAPE) are well known error metrics. The median log accuracy ratio (MdLQ) overcomes some of the drawbacks of the absolute percentage errors \cite{Morley2016} as the log-transformation tends to induce symmetry in positively skewed distributions, thus reducing bias. Interpreting MdLQ is not intuitive, a problem overcome by the median symmetric accuracy (MdSymA) which can be interpreted as a percentage error, similar to MAPE. Peak and total consumption errors can be calculated using the same formulae and are equivalent to the corresponding demand errors.

The consumption error measures are calculated for $N$, where $N$ are all $h_d^{(j)}$ assigned to $k_x$. 

\paragraph{Absolute Percentage Error}
\begin{eqnarray}
    mape &=& 100 \times \frac{1}{N} \sum_1^N {\frac{|d_d^{(j)}-d_x^{(R)}|}{d_d^{(j)}}} \\ 
    mdape &=& 100 \times median \bigg(\frac{|d_d^{(j)}-d_x^{(R)}|}{d_d^{(j)}}\bigg)
\end{eqnarray}
 
\paragraph{Median Log Accuracy ratio}
\begin{eqnarray}
    Q_d^{(j)} &=& \frac{d_x^{(R)}}{d_d^{(j)}}\\
    mdlq &=& median \big(log(Q_d^{(j)})\big)
\end{eqnarray}

\paragraph{Median Symmetric Accuracy}
\begin{equation}
    mdsyma =100 \times (\exp{(median \big(|log(Q_d^{(j)})|\big))} - 1)
\end{equation}

\subsection{Mean Peak Coincidence Ratio}
\label{mean_peak_coincidence_ratio}

For each daily load profile $l(t)_d$ the peaks are identified as all those values that are greater than half the maximum daily load profile value. The python package peakutils was used to extract the peak values and peak times for all daily load profiles and all representative daily load profiles. 

\begin{equation}
    PeakTimes_d^{(j)}, PeakValues_d^{(j)} = i, l(i)_d \\
\end{equation}

where $l(i)_d > 0.5 \times l(t)_d^{max}$ and $i=\{0, 1, ... 23\}$. The mean peak coincidence ratio for a single cluster is a value between 0 and 1 that represents the ratio of mean peak coincidence to the count of peaks in cluster $k_x$. The magnitude of the peak is not taken into account in calculating the mean peak coincidence ratio. The mean peak coincidence (denoted as MPC) was calculated from the intersection of the actual and cluster peak times for all $h_d^{(j)}$ assigned to $k_x$: 

\begin{equation}
    MPC_x = \frac{1}{\#h_d^{(j)}} \times \#\Big(PeakTimes_d^{(j)} \cap PeakTimes_x^{(R)}\Big) 
\end{equation}

\subsection{Entropy as a Measure of Cluster Specificity}
\label{entropy_measure_cluster_specificity}

Entropy H is used to quantify the specificity of clusters and is calculated as follows:

\begin{equation}
    H_x^{(f)} = - \sum_{i=1}^n {p(v_i) \log_2(p(v_i))}
\label{eq:entropy}
\end{equation}

Here $i = {1 ... n}$ are the values of a feature $f$ and $p(v_i)$ is the probability that daily load profiles with value $v_i$ for feature $f$ are assigned to cluster $k_x$. For example, $H_x^{(weekday)}$ expresses the specificity of a cluster with regards to day of the week, with $f=weekday$ and $i = \{Mon, Tues, Wed, Thurs, Fri, Sat, Sun\}$, where $p(Sun)$ is the likelihood that daily load profiles that are used on a Sunday are assigned to cluster $k_x$.

To calculate peak and total daily demand entropy, we created percentile demand bins. Thus the values of feature $f=peak\_demand$ are $i = \{0...99\}$ and $p(59)$ is the likelihood that daily load profiles with peak demand corresponding to that of the 60th peak demand percentile are assigned to cluster $k_x$.

%% file: sections/appendix_b.tex
\section{Clustering Experiments}
\label{appendixb}

We implemented our experiments in python 3.6.5 using k-means algorithms from scikit-learn (0.19.1) and self-organising maps from the SOMOCLU (1.7.5) libraries\footnote{The codebase is available online at https://github.com/wiebket/del\_clustering}.

Table \ref{tab:experiments} summarises the algorithms, parameters and pre-processing steps for each experiment, with $Zeros = True$ indicating that zero consumption values were retained in the input dataset. 

\begin{table}[h]
\begin{center}
{\caption{Experiment details}\label{tab:experiments}}
\scriptsize
\begin{tabular}{cllcc}
    \hline\noalign{\smallskip}
    \textbf{Exp.} & \textbf{Algorithm} & \textbf{Parameters} & \textbf{Pre-bin} & \textbf{Zeros} \\
    \noalign{\smallskip}\hline\noalign{\smallskip}
     1 & kmeans & $m\{5, 8, 11, ... 136\}$ & & True\\ \noalign{\smallskip}
     2 & kmeans & $m\{5, 8, 11, ... 136\}$ & & True\\
       & SOM & $s\{5, 7, 9, ... 29\}$ & & True\\
       & SOM+kmeans & $s\{30, 40, ... 90\}, m$ & & True\\ \noalign{\smallskip}
     3 & kmeans & $m\{5, 8, 11, ... 136\}$ & & False \\
       & SOM & $s\{5, 7, 9, ... 29\}$ & & False \\
       & SOM+kmeans & $s\{30, 40, ... 90\}, m$ & & False \\ \noalign{\smallskip}
     4 & kmeans & $m\{2, 3, ... 10\}$ & AMC & True\\
       & SOM & $s\{2, 3, 4, 5\}$ & AMC & True\\
       & SOM+kmeans & $s\{4, 7, 11, ... 20\}, m$ & AMC & True\\ \noalign{\smallskip}
     5 & kmeans & $m\{2, 3, ... 19\}$ & AMC & True\\
       & SOM+kmeans & $s\{4, 7, 11, ... 20\}, m$ & AMC & True\\ \noalign{\smallskip}
     6 & kmeans & $m\{2, 3, ... 19\}$ & AMC & False \\ \noalign{\smallskip}
     7 & kmeans & $m\{2, 3, ... 19\}$ & integral kmeans & True\\ \noalign{\smallskip}
     8 & kmeans & $m\{2, 3, ... 19\}$ & integral kmeans & False\\ \noalign{\smallskip}
\end{tabular}
\end{center}
\end{table}

\subsection{Clustering Algorithms}
An experiment run $i$ takes input array $X$ to produce cluster set $k^{(i)}$ and predict a cluster $k_x^{(i)}$ for each normalised daily load profile $n_d^{(j)}$ of household $j$ observed on day $d$. Variations of kmeans, self-organising maps (SOM) and a combination of the two algorithms were implemented to cluster $X$. The kmeans algorithm was initialised with a range of $m$ clusters. The SOM algorithm was initialised as a square map with dimensions $s_i \times s_i$ for $s_i$ in range $s$. Combining SOM and kmeans first creates a $s \times s$ map, which acts as a form of dimensionality reduction on $X$. For each $s$, kmeans then clusters the map into $m$ clusters. The mapping only makes sense if $s^2$ is greater than $m$. $m$ and $s$ are the algorithm parameters.

\subsection{Normalisation}
The table below lists the normalisation techniques applied.

\begin{table}[h]
\begin{center}
{\caption{Data normalisation algorithms and descriptions}\label{tab:normalisation_algorithms}}
\scriptsize
\begin{tabular}{p{0.17\linewidth} l p{0.45\linewidth}}
    \hline\noalign{\smallskip}
    \textbf{Normalisation} & \textbf{Equation} & \textbf{Comments} \\
    \noalign{\smallskip}\hline\noalign{\smallskip}
     Unit norm & $n_d^{(j)} = \frac{h_d^{(j)}}{|h_d^{(j)}|}$ & Scales input vectors individually to unit norm \\ \noalign{\smallskip}
     De-minning & $n_d^{(j)} = \frac{l(t)_d - l(t)_d^{min}}{|l(t)_d - l(t)_d^{min}|}$ & Subtracts daily min. demand from each hourly value, then divides each value by deminned daily total \footnote{proposed by \cite{Jin2017}} \\ \noalign{\smallskip}
     Zero-one & $n_d^{(j)} = \frac{h_d^{(j)}}{l(t)_d^{max}}$ & Scales all values to a range [0, 1]; retains profile shape but is very sensitive to outliers. \footnote{also known as min-max scaler}\\ \noalign{\smallskip}
     SA norm & $n_d^{(j)} = \frac{h_d^{(j)}}{\frac{1}{24} \times \sum_{t=0}^{23} {l(t)_d}}$ & Normalises all input vectors to mean of 1; retains profile shape but very sensitive to outliers. \footnote{introduced as a comparative measure, as it is frequently used by South African domain experts}\\ 
\end{tabular}
\end{center}
\end{table}

\subsection{Pre-binning}

\subsubsection{Pre-binning by average monthly consumption (AMC)} \mbox{}\\
To pre-bin by average monthly consumption, we selected 8 expert-approved bin ranges based on South African electricity tariff ranges. The average monthly consumption (AMC) for household $j$ over one year is:
\begin{equation}
AMC^{(j)}=\frac{1}{12} \sum_{month=1}^{12} \sum_{d=1}^{month_{end}} \sum_{t=0}^{23} 230 \times l(t)_d \textrm{ kWh}
\end{equation}
All the daily load profiles, $H^{(j)}$ of household $j$ were assigned to one of 8 consumption bins based on the value of $AMC^{(j)}$. Individual household identifiers were removed from $X$ after pre-binning.

\subsubsection{Pre-binning by integral k-means} \mbox{}\\
Pre-binning by integral k-means is a data-driven approach that draws on the work of \cite{Xu2017}. For the simple case where $t$ represents hourly values, pre-binning by integral k-means followed these steps:
\begin{enumerate}
    \item Construct a new sequence $c(t)$ from the cumulative sum of profile $n_d^{(j)}$ normalised with unit norm
    \item Append $l(t)_d^{max}$ to $c(t)$ -- this ensures that both peak demand and relative demand increase are taken into consideration
    \item Gather all features in array $X^C$ and remove individual household identifiers 
    \item Use the kmeans algorithm to cluster $X^C$ into $k=8$ bins, corresponding to the number of bins created for AMC pre-binning
\end{enumerate}

%% file: sections/appendix_c.tex
\section{Cluster Evaluation}
\label{appendixc}

\subsection{CI Score and Quantitative Results}

To ease the quantitative evaluation process and allow for comparison across metrics, Mean Index Adequacy (MIA), Davies-Bouldin Index (DBI) and the Silhouette Index were combined into a Combined Index (CI) score. $Ix$ is an interim score that computes the product of the DBI, MIA and inverse Silhouette Index. The CI is the log of the weighted sum of $Ix$ across all experiment bins. A lower CI is desirable and an indication of a better clustering structure. The logarithmic relationship between $Ix$ and the CI means that the CI is negative when $Ix$ is between 0 and 1, 0 when $Ix = 1$ and greater than 0 otherwise. For experiments with pre-binning, the experiment with the lowest $Ix$ score in each bin is selected, as it represents the best clustering structure for that bin. For experiments without pre-binning, $bins = 1$ and $N_{bin} = N_{total}$. Table \ref{tab:clusters_top10} shows the top ten experiments based on CI score.

\begin{table}[h]
{\caption{Top 10 runs ranked by CI score}\label{tab:clusters_top10}}
\scriptsize
\begin{center}
\begin{tabular}{cccccccccc}
    \hline\noalign{\smallskip}
    \textbf{\#} & \textbf{CI} & \textbf{DBI} & \textbf{MIA} & \textbf{Sil.} & \textbf{Exp.}& \textbf{Alg.}& \textbf{m} & \textbf{Norm.} \\
    \noalign{\smallskip}\hline\noalign{\smallskip}
1 & 2.282 & 2.125 & 0.438 & 0.095 & 2 & kmeans & 47 & unit \\ \noalign{\smallskip}
2 & 2.289 & 1.616 & 1.220 & 0.262 & 5 & kmeans & 17 & zero-one \\ \noalign{\smallskip}
3 & 2.296 & 1.616 & 1.220 & 0.260 & 4 & kmeans & 17 & zero-one \\ \noalign{\smallskip}
4 & 2.301 & 2.152 & 0.485 & 0.119 & 6 & kmeans & 82 & unit \\ \noalign{\smallskip}
5 & 2.316 & 2.115 & 0.447 & 0.093 & 2 & kmeans & 35 & unit \\ \noalign{\smallskip}
6 & 2.320 & 2.199 & 0.486 & 0.121 & 5 & kmeans & 71 & unit \\ \noalign{\smallskip}
7 & 2.349 & 2.152 & 0.481 & 0.143 & 7 & kmeans & 49 & unit \\ \noalign{\smallskip}
8 & 2.351 & 2.189 & 0.434 & 0.090 & 2 & kmeans & 50 & unit \\ \noalign{\smallskip}
9 & 2.354 & 2.111 & 0.476 & 0.128 & 8 & kmeans & 59 & unit \\ \noalign{\smallskip}
10 & 2.355 & 2.173 & 0.453 & 0.093 & 2 & kmeans & 32 & unit \\ \noalign{\smallskip}
\end{tabular}
\end{center}
\end{table}

\subsection{Experiments Ranked by Qualitative Score}

\begin{table}[ht!]
{\caption{Top runs ranked by qualitative scores}\label{tab:qual_ranked_results}}
\scriptsize
\begin{center}
\begin{tabular}{cccccc}
    \hline\noalign{\smallskip}
    \textbf{\#} & \textbf{Score} & \textbf{Exp.} & \textbf{Norm.} & \textbf{Pre-binning} & \textbf{Zeros}\\
    \noalign{\smallskip}\hline\noalign{\smallskip}
1 & 57.0 & 8 & unit & integral kmeans & False \\ \noalign{\smallskip}
2 & 65.0 & 5 & unit & AMC & True\\ \noalign{\smallskip}
3 & 117.5 & 6 & unit & AMC & False \\ \noalign{\smallskip}
4 & 143.5 & 7 & unit & integral kmeans & True\\ \noalign{\smallskip}
5 & 150.0 & 2 & unit & & True\\ \noalign{\smallskip}
6 & 205.0 & 5 & zero-one & AMC & True\\ \noalign{\smallskip}
7 & 208.0 & 4 & zero-one & AMC & True\\ \noalign{\smallskip}
\end{tabular}
\end{center}
\end{table}

\subsection{Comparison of Two Clustering Experiments}

\begin{figure}[!ht]
\centering
\begin{subfigure}{.49\textwidth}
  \includegraphics[width=\linewidth]{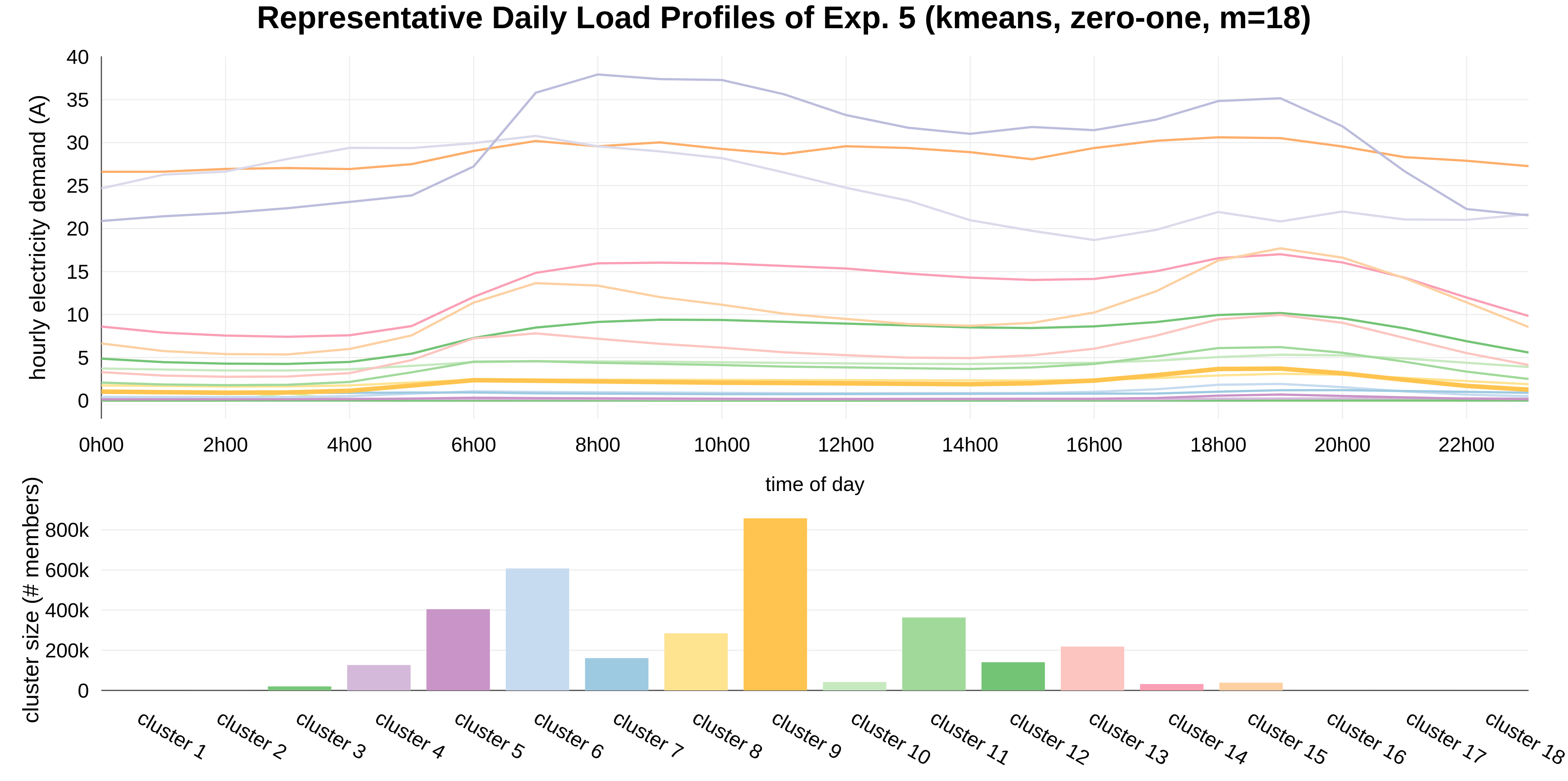}
\caption{RDLPs of exp. 5 (kmeans, zero-one)}
\label{fig:cluster_centroids_exp5_kmeans_zero-one}     
\end{subfigure}
\begin{subfigure}{.49\textwidth}
  \includegraphics[width=\linewidth]{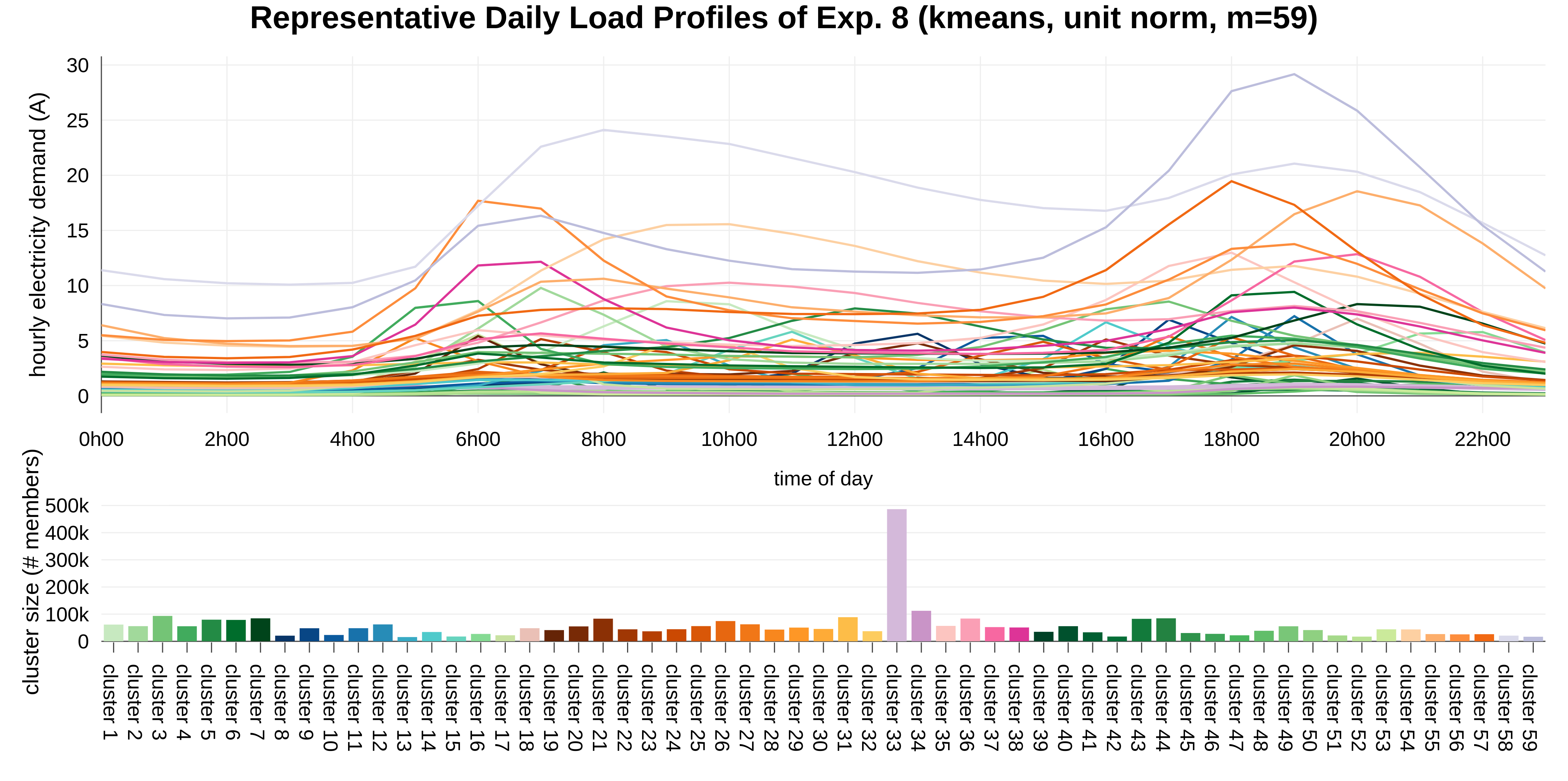}
\caption{RDLPs of exp. 8 (kmeans, unit norm)}
\label{fig:cluster_centroids_exp8_kmeans_unit_norm_bin}
\end{subfigure}
\caption{Comparison of RDLPs of clustering experiments}
\label{fig:cluster_comparison}
\end{figure}